\documentclass[manuscript,screen]{acmart}

\AtBeginDocument{%
  \providecommand\BibTeX{{%
    \normalfont B\kern-0.5em{\scshape i\kern-0.25em b}\kern-0.8em\TeX}}}

\usepackage{tabularx}
\usepackage{booktabs}
\usepackage{subcaption}
\usepackage{multirow}
\usepackage{pbox}
\usepackage{comment}
\usepackage{graphicx}
\usepackage{array}
\usepackage{tikz}
\usetikzlibrary{positioning}

\begin{document}

\title[Unveiling Social Media Comments with a Novel Named Entity Recognition System]{Unveiling Social Media Comments with a Novel Named Entity Recognition System for Identity Groups}

\author{Andres Carvallo}
\affiliation{%
\institution{CENIA}
 \country{Chile}}

\author{Tamara Quiroga}
\affiliation{%
 \institution{UC}
 \country{Chile}}

\author{Carlos Aspillaga}
\affiliation{%
  \institution{CENIA}
  \country{Chile}}

\author{Marcelo Mendoza}
\affiliation{%
\institution{UC}
 \country{Chile}}

\renewcommand{\shortauthors}{Carvallo et al.}

\begin{abstract}

While civilized users employ social media to stay informed and discuss daily occurrences, haters perceive these platforms as fertile ground for attacking groups and individuals. The prevailing approach to counter this phenomenon involves detecting such attacks by identifying toxic language. Effective platform measures aim to report haters and block their network access. In this context, employing hate speech detection methods aids in identifying these attacks amidst vast volumes of text, which are impossible for humans to analyze manually. In our study, we expand upon the usual hate speech detection methods, typically based on text classifiers, to develop a \textbf{Named Entity Recognition (NER) System for Identity Groups}. To achieve this, we created a dataset that allows extending a conventional NER to recognize identity groups. Consequently, our tool not only detects whether a sentence contains an attack but also tags the sentence tokens corresponding to the mentioned group. Results indicate that the model performs competitively in identifying groups with an average f1-score of 0.75,  outperforming in identifying ethnicity attack spans with an f1-score of 0.80 compared to other identity groups. Moreover, the tool shows an outstanding generalization capability to minority classes concerning sexual orientation and gender, achieving an f1-score of 0.77 and 0.72, respectively. We tested the utility of our tool in a case study on social media, annotating and comparing comments from Facebook related to news mentioning identity groups. The case study reveals differences in the types of attacks recorded, effectively detecting named entities related to the categories of the analyzed news articles. Entities are accurately tagged within their categories, with a negligible error rate for inter-category tagging.

\end{abstract}



\maketitle

\section{Introduction}

Social media platforms have attracted millions of users who gather information and comment on daily occurrences. The adoption of social media as the primary media of human interaction has become so deep that these platforms can now influence public opinion and generate polarizing dynamics that impact society. In this high-influence landscape, bad actors have found fertile ground to spread ideas and hate messages, targeting individuals and groups through the use of hate speech.

To counteract the harmful effects of toxic language, researchers and practitioners have developed methods for automatically detecting hate speech ~\cite{macavaney2019hate,yin2021towards,poletto2021resources}. Predominantly, these methods rely on text classifiers that determine whether a sentence contains an attack. Enhancements to these models include multiclass classifiers, which additionally identify if the sentence targets a specific group. Although their effectiveness has been questioned, revealing limitations in transfer learning scenarios, these models currently serve as the primary automatic tool for analyzing vast amounts of text on social media.

In this study, we extend the conventional approach of hate speech detection, typically based on classification, by developing a Named Entity Recognition (NER) system to tag identity groups in text . The creation of a NER tool for annotating identity groups merges two approaches: the use of toxic language against a specific group, generally addressed from a multiclass classification perspective, and the tagging of enclosed text indicating a group, approached using NER. By integrating these two approaches, we have developed a system capable of tagging tokens that refer to a specific group.

The advantages of this approach over a traditional text classifier include not only the ability to detect toxic language in a sentence but also to pinpoint specific tokens related to the group. Furthermore, our tool directly links the attack to a particular identity group, with the tagged tokens referring to the identity group. This task is impossible with standard text classifiers, as they only provide annotations at the sentence level. On the other hand, our tool enhances what conventional NER systems can do. Typical NER tools are not trained to identify identity groups; they usually identify entities with broader granularity, such as person, organization, and location, among other categories. By expanding NER to include identity groups, our tool enables the annotation of large volumes of text, facilitating the analysis of threats to these groups. Another advantage of our tool is its ability to identify references to one or more groups within a sentence, facilitating intersectional analysis of categories.

Attacks can be executed in either a coupled or decoupled manner relative to the mention. A decoupled attack establishes an offensive relation to the mentioned entity. For example, the phrase 'inflame black people' identifies the group with the token 'black people,' while the toxicity is conveyed through the token 'inflame.' In contrast, a coupled attack overlays the attack and the mention at the lexical level simultaneously through what is termed an 'offensive mention.' An example of an 'offensive mention' is the term 'n*r', which refers to black people using toxic language. Our tool is capable of detecting both cases, including offensive mentions and decoupled attacks from the mention. In both scenarios, the specified group is identified through a decoupled mention from the attack or an offensive reference.

To address this challenge, we have merged two tasks, multiclass classification, and NER, by aligning two datasets: HateNorm~\cite{Masud:22}, which focuses on text spans, and Jigsaw Toxicity~\cite{Sorensen:17}, the dataset used for the toxic comment classification challenge. To align these datasets, we propose a methodology for selecting sentences tagged using a NER trained on HateNorm and sentences annotated at the span-level based on a multiclass classifier developed from the Toxicity dataset. Since the Toxicity dataset is annotated at the sentence level, we apply the NER trained on HateNorm to detect groups and then classify the tagged spans using toxicity. If the annotation from toxicity matches the class predicted for the span text by the multiclass classifier, we annotate the short span in the detected class. Through this process, we align the Toxicity dataset with the tokens detected by the HateNorm NER, creating a dataset with text spans marked in specific identity groups. Subsequently, this dataset was utilized to fine-tune a conventional NER by adding identity groups to the detected entities. Experimental results show that the NER for identity groups performs well on test data.

We developed a case study to illustrate the utility of our tool. The study is based on a dataset of news articles downloaded from ABC News, a source that categorizes news mentioning identity groups. Subsequently, we downloaded comments triggered by these news articles on Facebook. By utilizing our tool to annotate these comments, we describe differences in terms of the types of attacks and groups mentioned. The case study reveals differences in the types of attacks recorded, effectively detecting named entities related to the categories of the analyzed news articles. Entities are accurately tagged within their categories, with a negligible error rate for inter-category tagging.

The main contributions of the paper are:

\begin{itemize}
    \item[--] We align two datasets, Jigsaw toxicity, and HateNorm, to generate a dataset for training a Named Entity Recognition (NER) system that can detect mentions of identity groups.
    \item[--] We apply the NER for identity groups to a case study comparing news across two social media platforms. The case reveals differences in the types of attacks and groups mentioned.  
    
\end{itemize}

The paper is structured as follows. Section \ref{relwork} introduces the related work. Section \ref{ner} describes the methodology employed in developing the NER (Named Entity Recognition) System for identity Groups. Section \ref{case} introduces the case study. Finally, Section \ref{conc} offers concluding remarks and outlines future work. 

\section{Related work}
\label{relwork}

In recent years, the field of detecting toxic comments and identifying specific text spans related to these comments has gained significant academic interest. The evolution from traditional methods to more advanced techniques, particularly the use of Large Language Models (LLMs), marks a pivotal shift in tackling online hate speech and offensive content ~\cite{alatawi2021detecting,gamback2017using,saleh2023detection,wei2021offensive,zhang2018detecting,del2017hate}. Early research efforts, such as those by ElSherief et al. and Zampieri et al, laid foundational groundwork by focusing on linguistic analyses and the prediction of offensive post types in social media, offering critical insights into the nature of hate speech and its propagation online \cite{elsherief2018hate, zampieri2019predicting}. These studies underscored the importance of not only detecting toxic content but also understanding its targets and nuances, paving the way for more targeted approaches in subsequent research.

Subsequent works, including Shvets et al. and initiatives like the OSACT5 and HASOC shared tasks in 2022, expanded on these foundations by exploring specific aspects of social media hate speech, such as target identification and the detection of offensive language across various languages and regions \cite{shvets2021targets, mubarak2022overview, satapara2022overview}. These efforts highlight the growing complexity of hate speech detection and the need for models that can navigate the nuances of language and cultural context effectively.

Despite these advancements, a gap remains in the targeted identification within specific toxic text spans, a crucial aspect for understanding and mitigating the impact of online hate. This gap underscores the need for methodologies that go beyond detection to dissect the elements of toxic comments, including the identification of targeted identity groups ~\cite{goodall2013challenging,ștefuanițua2021hate}.

Our research addresses this need by leveraging Named Entity Recognition (NER) models to detect and identify the targeted identity group within toxic text spans. The application of NER and NLP techniques in fields such as medicine and education \cite{alvarez2021automatic, carvallo2023automatic, munoz2023linkmed, carvallo2020automatic, carvallo2019comparing, carvallo2020neural} supports the feasibility and utility of this technology in our approach. Building upon the foundational work of earlier studies and the innovative use of LLMs for toxic comment detection ~\cite{li2023hot,mishra2023exploring}, our dataset and model approach represent a significant contribution to the field. By focusing on identity categories such as gender, sexual orientation, ethnicity, and religion, we provide a comprehensive tool for analyzing and combating hate speech in a nuanced manner. This contribution is particularly relevant in light of the collaborative efforts by the Conversation AI team with Jigsaw and Google on the \textbf{Jigsaw Toxicity Comment Classification dataset} and the \textbf{HateNorm 2023 shared task}, which emphasize the critical need for nuanced, identity-specific analysis in toxic commentary detection ~\cite{Sorensen:17,Masud:22}.

In summary, our work not only builds upon the significant efforts of previous studies but also introduces an innovative approach to identifying targeted identity groups within toxic text spans. By integrating insights from the proposed papers and aligning them with the advancements in NER and LLM technologies, we aim to further the field's understanding and capability in addressing the multifaceted nature of online hate speech.

\section{NER for identity Groups}
\label{ner}

Our proposed tool introduces a specialized Named Entity Recognition (NER) dataset aimed at training models to identify text spans featuring attacks against critical entity groups such as ethnicity, sexual orientation, gender, and religion. This dataset is uniquely structured to capture the nuances of language in these contexts, enabling the model to accurately detect forms of hate speech and discrimination in specific text spans in a comment. 

\begin{figure}[h]
    \centering
    \includegraphics[width=\linewidth]{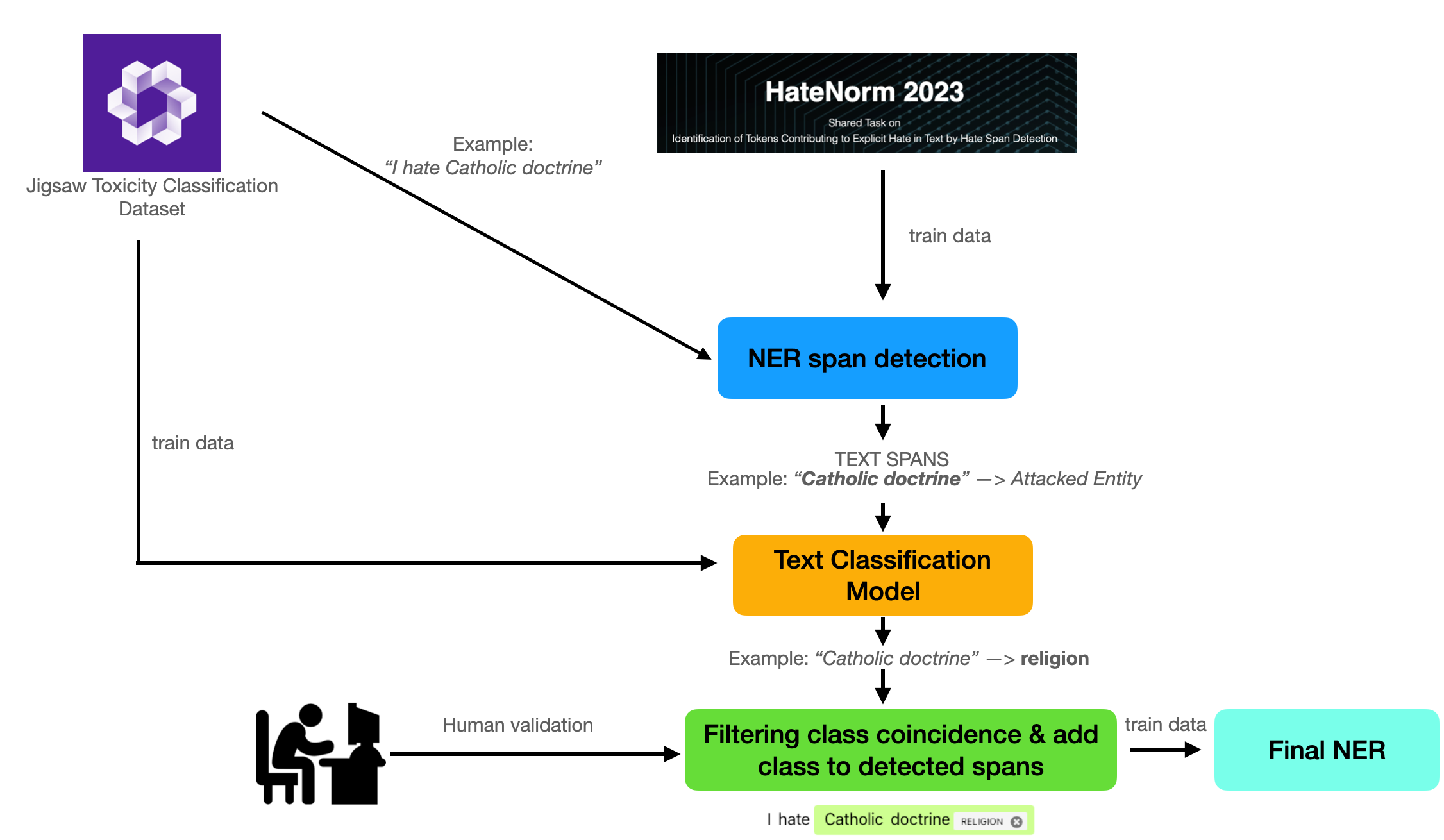}
    \caption{Pipeline Diagram for Classification, NER, and Identification of Text Spans Attacking Entity Groups}
    \label{fig:pipeline-diagram}
\end{figure}

Figure \ref{fig:pipeline-diagram} shows our methodology for generating the NER dataset, which is later used for training the final NER model. It starts with the Jigsaw toxicity dataset, primarily used for classification that consists of annotated texts with labels indicating targets of attacks. Although it does not provide specific text spans where these attacks occur, the Jigsaw dataset ~\cite{Sorensen:17} is the cornerstone of our entity extraction process, paving the way for the subsequent application of the trained NER model. 

Our process begins with developing a transformer-based language model augmented with a classification layer. This model was designed to predict the category of the entity group targeted in a given text. Moreover, we further refined our approach by training a Named Entity Recognition (NER) model, incorporating a language model fitted with a token classification layer, and utilizing the HateNorm dataset~\cite{Masud:22}. Unlike the Jigsaw dataset, HateNorm provides precise span annotations of mentions within texts but omits the classification of the targeted entity groups.

We then applied the trained NER model to the Jigsaw toxicity dataset for entity extraction. The extracted entities were further processed through our classification model \textbf{to select instances where the extracted text class aligns with the provided ground truth.} Accordingly, the dataset is composed solely of examples where the predicted class by the toxicity classifier, based on tokens identified by HateNorm's Named Entity Recognition (NER), aligns with the human annotation provided at the sentence level by the Jigsaw toxicity dataset. Consequently, the dataset includes examples that match both the automatic annotation on the span and the human annotation at the sentence level. Therefore, under these conditions, the dataset is aligned in both annotations, ensuring that the examples have consistent annotations.

The final step of our methodology is an exhaustive manual review and correction phase. This phase is indispensable in validating the entities identified by the HateNorm-based model. It ensures that the entities align accurately with the intended groups, eliminates irrelevant terms associated with the attacked entities, and verifies the contextual relevance of the entities within the attack narrative. This rigorous process is critical to extracting entities specifically within the context of an attack, thereby significantly enhancing the reliability of the final NER model.

Consequently, the resultant dataset was employed to train a NER model, thoroughly cleaning and integrating both text spans and their respective targeted groups. This model mirrors the architecture of the initial language model, featuring a token classification layer.

Our integrated approach serves a dual purpose: it accurately pinpoints where in the text an attack is directed at a specific entity (such as race, ethnicity, gender, or sexual orientation) and identifies the specific entity being targeted. This dual functionality enables a more in-depth and nuanced analysis of the textual data, providing a comprehensive understanding of the context and nature of the attack. It is important to note that our analysis is tailored explicitly to scenarios where an entity group is the subject of the attack. 

\begin{figure}[h]
    \centering

    \begin{subfigure}{\linewidth}
        \centering
        \fbox{\includegraphics[width=0.85\linewidth]{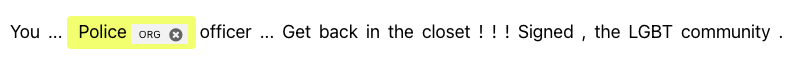}}
        \caption{Traditional NER Annotations}
        \label{fig:ner_example1}
    \end{subfigure}

    \begin{subfigure}{\linewidth}
        \centering
        \fbox{\includegraphics[width=0.85\linewidth]{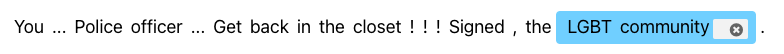}}
        \caption{HateNorm NER Annotations}
        \label{fig:ner_example2}
    \end{subfigure}

    \begin{subfigure}{\linewidth}
        \centering
        \fbox{\includegraphics[width=0.85\linewidth]{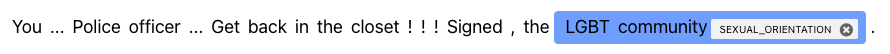}}
        \caption{ NER Annotations obtained using our strategy}
        \label{fig:ner_example3}
    \end{subfigure}

    \caption{Comparative NER Annotations Identifying Attacked Entities and Their Text Spans}
    \label{fig:ner_examples}
\end{figure}

As seen in Figure \ref{fig:ner_examples}, we show three approaches to NER annotations, each representing a different depth of analysis. The first image illustrates traditional NER annotations, which involve identifying general knowledge entities such as organizations, locations, persons, or dates. The second example, from the HateNorm dataset, marks a significant shift towards domain-specific knowledge. Here, the annotations are nuanced, identifying specific text spans where an entity group is attacked. However, this method lacks specifying which entity group is under attack. The third example, representing our work, elevates the annotation process to a new level. It identifies the specific spans of text where attacks occur and distinctly annotates which entity group is being targeted, namely religion, ethnicity, sexual orientation, or gender.

\begin{table}
\centering
\caption{Distribution of Entity Groups Across Dataset Partitions (Train, Validation, Testing, and Total)}
\begin{tabular}{lcccccc}
\toprule
             & Total & Religion & Ethnicity & Sexual Orientation & Gender \\
\midrule
Train        &   72,678    &   36,909         &   26,807        &  4,650      &   4,312                 \\
Validation   &   3,970    &  1,954          &  1,553         &  250      &    213                \\
Testing      &   4,190    &    2,139        &   1,477        &   244    &    230                \\
\midrule
Total        &  80,838   &  41,002  &  29,837  &  5,144  &  4,755  \\
\bottomrule
\end{tabular}

\label{tab:entity-distribution}
\end{table}

As seen in Table \ref{tab:entity-distribution}, the dataset employed for the NER task is segmented into three distinct subsets: training, validation, and testing. The training set, which serves as the foundation for model learning, is the most extensive, encompassing 72678 entities. Within this subset, the distribution of entities spans various categories, with Religion being the most represented at 36909 entities. Ethnicity follows, with a significant count of 26807 entities. In contrast, Sexual Orientation and Gender entities are less prevalent, totaling 4650 and 4312, respectively. The validation set, crucial for fine-tuning the model, includes a more modest total of 3970 entities. As the training set, Religion and Ethnicity dominate this subset, accounting for 1954 and 1553 entities, respectively. Entities representing Sexual Orientation and Gender are comparatively fewer, amounting to 250 and 213, respectively. In the testing set, which is critical for evaluating the model's performance, there are 4190 entities. Here again, Religion and Ethnicity are the most represented categories, with 2139 and 1477 entities, respectively. Sexual Orientation and Gender maintain their trend of lower representation with 244 and 230 entities, respectively. When aggregating the totals across all subsets, the dataset comprises 80,838 entities. The category-wise aggregation reveals a consistent trend across the subsets, with Religion entities being the most numerous at 41002, followed by Ethnicity at 29837. Sexual Orientation and Gender entities, while essential, are less represented, with totals of 5144 and 4755, respectively. This distribution underscores the dataset's emphasis on Religion and Ethnicity entities and highlights the challenge of achieving a balanced representation of different entity types in NER tasks.

\begin{table}[h]
\centering
\caption{Sample of Most Frequent Words in Spans for Each identity Group}
\label{tab:entity_words}
\begin{tabularx}{\textwidth}{|>{\centering\arraybackslash}X|X|}
\hline
\textbf{Identity Group} & \textbf{Words Sample} \\
\hline
Gender & women, male, men, transgender, feminism, transgendered, gender, man, woman, trans \\
\hline
Ethnicity & white, black, immigrants, non-white, racist, racism, asian, racial, african, latino \\
\hline
Sexual Orientation & gay, lesbian, bisexual, LGBT, homosexual, homosexuality, same-sex, anti-gay, heterosexuals, homophobia \\
\hline
Religion & islam, catholic, muslim, jews, jewish, terrorism, islamophobia, palestinians, extremists, atheists  \\
\hline
\end{tabularx}
\end{table}

As shown in Table \ref{tab:entity_words}, the table offers an insightful sample of the most frequent words associated with each identity group. This selection of words reflects an in-depth understanding of the context and nuances of each group, showcasing not just directly related terms but also those indicative of attacks or discrimination. For instance, in the context of the sexual orientation entity group, words such as "homophobia" clearly demonstrate this dual relevance. They are inherently related to the group and carry connotations of the attacks these entities might face. This careful curation of words for each entity group - gender, ethnicity, sexual orientation, and religion - enhances the model's ability to accurately identify and classify instances of attacks or discrimination in text. Notably, the chosen words encapsulate the group's identity and the nature of the challenges or biases they can encounter, providing a substantial foundation for the NER model's training and subsequent performance.

\begin{table}[h]
\centering
\caption{Performance Metrics of the NER Model}
\label{tab:ner_model_performance}
\begin{tabular}{@{}l|lll|l@{}}
\toprule
Entity Group        & Precision & Recall & F1-Score & Support \\ 
\midrule
Religion            & 0.76      & 0.69   & 0.72     & 2139    \\
Ethnicity           & 0.80      & 0.80   & 0.80     & 1477    \\
Sexual Orientation  & 0.80      & 0.75   & 0.77     & 244     \\
Gender              & 0.72      & 0.73   & 0.73     & 230     \\
\midrule
Micro Avg           & 0.77      & 0.74   & 0.75     & 4090    \\
Macro Avg           & 0.77      & 0.74   & 0.76     & 4090    \\
Weighted Avg        & 0.77      & 0.74   & 0.75     & 4090    \\
\bottomrule
\end{tabular}
\end{table}

The performance metrics shown in Table \ref{tab:ner_model_performance} reveal the final NER model's proficiency,  mainly for supporting class imbalance within the dataset. This model can generalize across classes within varying representation levels. For instance, the model achieves a high f1-score of 0.80 in the \textit{'Ethnicity'} category, which suggests strong performance despite potential data skewness. This is particularly noteworthy given the relatively lower number of examples in the \textit{'Sexual Orientation'} category (244 instances), where the model still manages an f1-score of 0.77, indicating its effective handling of class imbalance.
The \textit{'Religion'} category, with the highest support of 2139 instances, obtains a solid f1-score of 0.72, further reinforcing the model's capability to maintain accuracy across different scales of data representation. Similarly, the \textit{'Gender'} category, despite having the most minor support (230 instances), shows a comparable f1-score of 0.73. This consistency across categories with varied instance counts highlights the model's robustness.
Moreover, the balanced performance across micro, macro, and weighted averages (f1-scores of 0.75, 0.76, and 0.75, respectively) suggests the model is not overly biased towards any particular class, a critical factor in scenarios with imbalanced data. This adaptability is crucial for real-world applications where data representation might not be uniform.
The comprehensive approach to model evaluation, leveraging a mix of heuristic-based annotations, algorithmic classifications, and human validation, ensures that the model is usable for practical deployment. Its ability to accurately interpret and analyze text across a spectrum of scenarios, as evidenced by the metrics, demonstrates its potential for complex text analysis tasks where accuracy, sensitivity, and the ability to handle class imbalance are critical.

\section{Case study}
\label{case}

\subsection{Methodology and main findings}

The study case in which we applied our tool consists of a collection of news articles along with their corresponding comments on Facebook. We utilized ABC News as our information source, a network known for categorizing news items according to the identity groups mentioned in either the headline or body of the article. 

\begin{table}[ht!]
    \centering
    \caption{News analyzed in the case study.}
    \resizebox{0.88\columnwidth}{!}{
    \begin{tabular}{c|c|l|c|c|c|c} \hline
                    & ID & Headline    & Date  & Comments  & Shares  & Reactions \\  \hline \hline \vspace{0.1cm}
    \multirow{12}{*}{\rotatebox[origin=c]{90}{Gender}} 
    & (1) & \pbox{7.1cm}{The Ohio Senate voted to pass a bill that would limit gender-affirming care for transgender minors}     &  12/13/23     &   383        &  36       &   596        \\ \cline{3-3} \vspace{0.1cm}
    & (2) & \pbox{7.1cm}{Australia's gender pay gap has fallen to a new low of 21.7 per cent. Do you think the gender pay gap is closing?} &  11/28/23     &  27         &  1       & 60           \\ \cline{3-3} \vspace{0.1cm}
    & (3) & \pbox{7.1cm}{"It's boss as, girls getting around with forklifts and [doing] something males always do," Tahlia Quinlivan said.} &  01/15/24     &  159         & 11        &  251         \\ \cline{3-3} \vspace{0.1cm}
    & (4) &  \pbox{7.1cm}{Gender-affirming hormone therapy improves the mental health of transgender adolescents and teenagers, a new study released Wednesday in the New England Journal of Medicine showed.} &  01/19/23     &  258         & 13        &  712         \\ \cline{3-3} \vspace{0.1cm}
    & (5) & \pbox{7.1cm}{In almost six years as leader, Jacinda Ardern faced the aftermath of a massacre and delivered stinging rebukes to those questioning her age and gender. Here's a look back at her time in office.} & 01/19/23 & 35          & 5        & 205 \\ \hline \vspace{0.1cm}
    \multirow{11}{*}{\rotatebox[origin=c]{90}{Ethnicity}} 
    & (6) & \pbox{7.1cm}{Germany on Tuesday banned the neo-Nazi group Hammerskins Germany and raided homes of dozens of its members. The group is an offshoot of an American ring-wing extremist group and plays a prominent role across Europe.}     &  09/19/23     & 159          & 33        & 409          \\ \cline{3-3} \vspace{0.1cm}
    & (7) & \pbox{7.1cm}{An upstate New York museum is featuring homemade dolls depicting African American life as an homage to their makers and as a jumping off point into the history of oppression faced by the Black community.}     &  10/22/23     &   103        &    8     &   185        \\ \cline{3-3} \vspace{0.1cm}
    & (8) & \pbox{7.1cm}{Supreme Court rejects appeal of Derek Chauvin conviction in the killing og George Floyd}     &  11/20/23     &     733      & 72        & 1127          \\ \cline{3-3} \vspace{0.1cm}
    & (9) & \pbox{7.1cm}{King Charles, Princess Kate identified by British press as royals named in Dutch version of "Endgame" amid racial controversy}     &  12/01/23     &  62         & 5        & 80          \\ \cline{3-3} \vspace{0.1cm}
    & (10)& \pbox{7.1cm}{Nazi flags to be banned under new Queensland hate symbol laws.}     &  10/12/23     &   36        &  4       &  250         \\ \hline \vspace{0.1cm}
    \multirow{9}{*}{\rotatebox[origin=c]{90}{Sexual orientation}} 
    & (11) & \pbox{7.1cm}{The Florida Department of Education has effectively banned AP Psychology}     &  08/03/23     &  222         & 24        & 208          \\ \cline{3-3} \vspace{0.1cm}
    & (12) & \pbox{7.1cm}{Hate crimes in the U.S. remained at the same level in 2021 as they were in 2020, FBI says, even as hate crimes based on a person's sexual orientation increased}     &  12/12/22     &  154         & 16        & 122          \\ \cline{3-3} \vspace{0.1cm}
    & (13) & \pbox{7.1cm}{A new bill proposed in one state would take Florida's controversial "Don't say Gay" law even further}     &  02/08/23     &  305         & 21        & 265          \\ \cline{3-3} \vspace{0.1cm}
    & (14) & \pbox{7.1cm}{FDA drops blood donation restrictions specific to gay and bisexual men} &  05/14/23  &  122         & 18        &  147         \\ \cline{3-3} \vspace{0.1cm}
    & (15) & \pbox{7.1cm}{The Florida Board of Education votes to expand so-called "Don't Say Gay" rules through 12th grade}     &  04/19/23     &  398         & 63        & 400          \\ \hline \vspace{0.1cm}
    \multirow{8}{*}{\rotatebox[origin=c]{90}{Religion}} 
    & (16) & \pbox{7.1cm}{Ramaswamy says he's not "pastor in chief" as GOP voters question his religion}     &  09/14/23     &  101         & 13        & 107          \\ \cline{3-3} \vspace{0.1cm}
    & (17) & \pbox{7.1cm}{The Israel-Hamas war has college campuses on edge. How some are tackling the issue}     &  11/29/23     & 190          & 4        &    71       \\ \cline{3-3} \vspace{0.1cm}
    & (18) & \pbox{7.1cm}{DeSantis, Trump court Iowa's evangelical voters, promising Christian-focused policy}     &  12/28/23     &  184         & 4        &  181         \\ \cline{3-3} \vspace{0.1cm}
    & (19) & \pbox{7.1cm}{Gen Z gym bros resurrecting Christianity as religion makes godlike gains on social media}     &  06/08/23     & 77          & 12        &  401         \\ \cline{3-3} \vspace{0.1cm}
    & (20) & \pbox{7.1cm}{Islam is the world's fastest growing religion due to migration and high birth rates}     &  06/28/23     & 28          & 4        & 55          \\ \hline \hline \vspace{0.1cm}
    & Total &      &       & 3736          & 367        & 5832          \\ \hline
    \end{tabular}
    }
    \label{tab:news}
\end{table}

This organization of news by groups facilitated our search for content likely to generate related comments on social media concerning these groups. To compile the set of news articles for analysis, we employed the META Crowdtangle search tool, enabling access to content from one or more press outlets with public pages on Facebook. In the case of ABC News, we focused on their three most followed Facebook pages: \url{https://www.facebook.com/ABCNews} (18 million users), \url{https://www.facebook.com/abcnews.au} (4.7 million users), and \url{https://www.facebook.com/WorldNewsTonight} (5 million users). We utilized Crowdtangle's search engine on the specified sources to assemble our news collection. The keywords used for different categories were \textit{'gender'} for the \textit{Gender} category, \textit{'racism ethnicities'} for \textit{Ethnicity}, \textit{'sexual orientation'} for \textit{Sexual Orientation}, and \textit{'religion'} for \textit{Religion}. Searches were parameterized over a one-year observation window. The first five news articles were collected for each category, resulting in 20 news articles for the case study. The news corpus comprises 3,736 comments, 367 shares, and 5,832 reactions. Details of the constructed corpus are provided in Table \ref{tab:news}.

We applied our tool to each comment of every news article in our corpus. Table \ref{tab:mentions}, alongside the ID of each news item, displays the total number of comments in which the tool tagged mentions about any of the four analyzed categories. The 'intersections' column indicates the count of co-occurrences of two or more mentions in the same comment. For instance, for news item (1), the intersection (G,G,26) is displayed, signifying 26 co-occurrences of mentions of two or more entities tagged in the 'gender' category. The last column of Table \ref{tab:mentions} provides examples of entity mentions identified by our tool.

\begin{table}[h!]
    \centering
    \caption{Mentions per category.}
    \begin{tabular}{c|c|c|c|c|l|l} \hline
ID   & Gender  &  Ethnicity  & Sexual Or.   &    Religion  &  Intersections       &  Examples           \\ \hline \hline
(1)  &  54     &  2          & 3            &    0         &  (G,G,26)            &  transgender, trans, women  \\    
(2)  &   1     &  0          & 0            &    0         &  -                   &  pays women \\
(3)  &  43     &  0          & 0            &    0         &  (G,G,12)            &  women, men \\
(4)  &   8     &  0          & 0            &    0         &  (G,G,2)             &  transgender \\
(5)  &   3     &  1          & 0            &    0         &  (G,G,1)             &  female misogyny, women \\ \hline
(6)  &   0     &  3          & 0            &    0         &  -                   &  white \\
(7)  &  26     & 18          & 0            &   11         &  (G,G,73),(R,R,12)   &  Hamas, Hamas bombed, Jews \\
(8)  &   6     &  5          & 0            &    4         &  (R,R,3)             &  racists, Pakistan, man \\
(9)  &   0     &  1          & 0            &    0         &  -                   &  white \\
(10) &   0     &  0          & 0            &    1         &  -                   &  Hamas \\ \hline
(11) &   2     &  0          & 8            &    7         &  (S,S,7),(R,R,2)     &  globalist, gay mafia, christian \\
(12) &   0     &  1          & 0            &    0         &  -                   &  white christian nationalism \\
(13) &   1     &  0          & 17           &    4         &  (S,S,6),(R,R,1)     &  gay, tansgender, facist christians \\
(14) &   0     &  2          & 8            &    0         &  (S,S,1)             &  gay, white \\
(15) &   1     &  4          & 60           &    3         &  (S,S,22),(R,R,3)    &  gay, homosexuality, homosexual \\ \hline
(16) &   0     &  1          & 0            &    2         &  -                   &  religion, Jewish, white \\
(17) &   0     &  0          & 0            &    7         &  (R,R,6)             &  Allah, Israelis, murderers rapist \\
(18) &   0     &  0          & 0            &   22         &  (R,R,14)            &  christian, religion \\
(19) &   0     &  0          & 0            &    5         &  (R,R,2)             &  religion, christianity \\
(20) &   0     &  0          & 0            &   14         &  (R,R,20)            &  Islam \\ \hline \hline
Total & 145    & 38          & 96           &   80         &  -                   &  -   \\ \hline
 
    \end{tabular}
    \label{tab:mentions}
\end{table}

Table \ref{tab:mentions} reveals that users' most frequently used category in these news items is 'gender,' with 145 comments. This is followed by 'sexual orientation' with 96 comments, 'religion' with 80, and 'ethnicity' with 38 comments. Regarding intersections, the results indicate a dominance of intra-category intersections. Examples of such intersections include comments in news item (1), featuring 26 intra-category intersections in 'gender', and news item (7) with 73 intra-category intersections in 'gender'. In news item (1), intra-category intersections are exemplified by overlaps between 'biological men' and 'transgender', and 'biological men' and 'outperform women'. In news item (7), examples include intersections between 'Hamas' and 'kill babies', and 'Muslims' and 'Zionist'. Additionally, Table \ref{tab:mentions} also reveals inter-category intersections in four news items: intersections occur between 'gender' and 'religion' in the news item (7) and between 'sexual orientation' and 'religion' in news items (11), (13), and (15). Overall, a low level of inter-category intersectionality is observed.

We examined the linear dependency between interaction variables (comments, shares, and reactions) and mentions of identity group categories. The Pearson linear correlation coefficients are presented in Table \ref{tab:correlations}.

\begin{table}[h!]
    \centering
    \caption{Correlation between interaction variables and mentions.}
    \begin{tabular}{l|c|c|c|c|c|c|c} \hline
& Comments	            &  Shares	& Reactions	& Gender	& Ethnicity	    & Sexual Or. & Religion \\ \hline
Comments	            &  1.000	& 0.880	    & 0.811	    & 0.219	        & 0.141	& 0.340	 & -0.027  \\
Shares	                &  -	& 1.000	    & 0.737	    & 0.114	        & 0.203	& \textbf{0.557}	 & -0.180  \\
Reactions	            &  -	& -	    & 1.000	    & 0.261	        & 0.141	& 0.057	 & -0.181  \\
Gender	                &  -	& -	    & -	    & 1.000	        & 0.284	& -0.110 & -0.145  \\
Ethnicity	            &  -	& -	    & -	    & -	        & 1.000	& 0.075	 & 0.177   \\
Sexual Orientation	    &  -	& -	    & -	    & -	    & -	& 1.000	 & -0.053  \\
Religion	            &  -	& - 	& -	    & -	    & -	& - & 1.000   \\ \hline
    \end{tabular}
    \label{tab:correlations}
\end{table}

Table \ref{tab:correlations} reveals a low linear correlation between interaction variables and identity group categories. A linear dependency is observed only between the 'shares' interaction variable and mentions of groups in the 'sexual orientation' category, with a correlation of 0.557.

\vspace{2mm}

\noindent \textbf{Main findings}. The main findings of the case study suggest a predominance of comments that involve mentions of 'gender' and 'sexual orientation', with a high level of intra-category intersectionality and low inter-category intersectionality. The study also reveals a weak correlation between interaction variables and mentions of groups, with the only significant linear dependence being between 'shares' and mentions of 'sexual orientation'.

\subsection{Inspection of relevant cases}

We examined several examples to illustrate the findings our tool can uncover. Three news articles were selected based on the volume of comments featuring mentions of identity groups detected by our tool. These are article (1) from the 'gender' category, article (7) from the 'ethnicities' category, and article (15) from the 'sexual orientation' category.

\begin{figure}[h!]
    \centering
    \includegraphics[width=15cm]{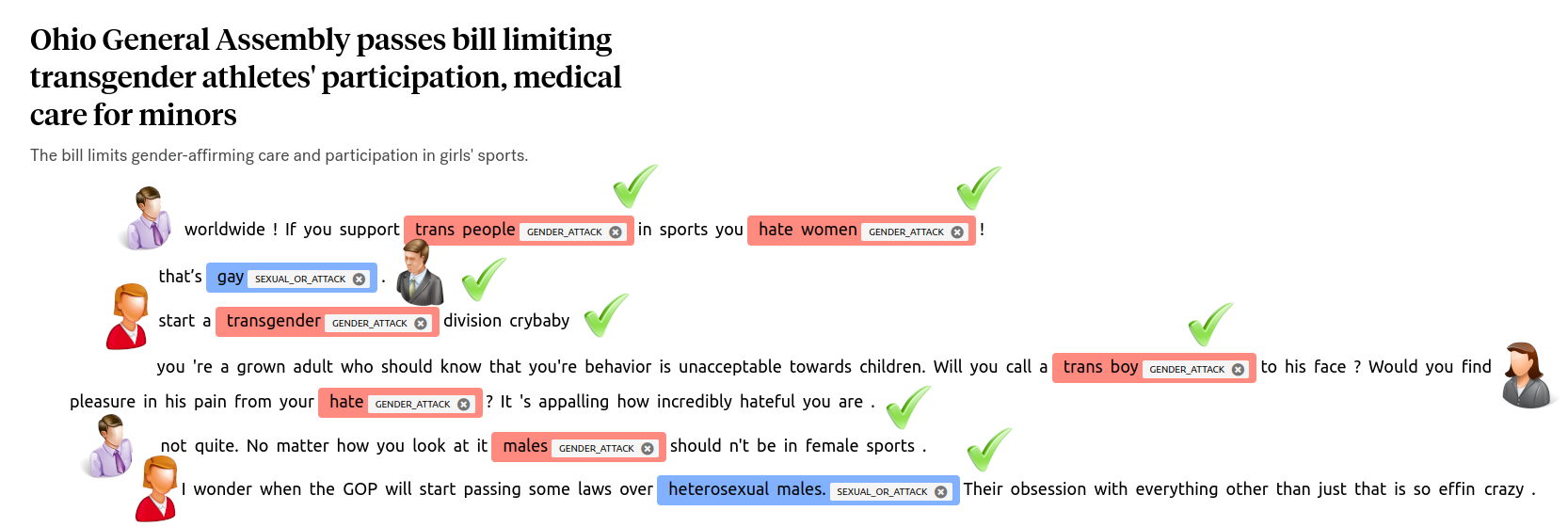}
    \caption{News (1) showing interactions between users. Our tool detects some gender and sexual orientation attacks during the conversation.}
    \label{fig:case_1}
\end{figure}

Starting with article (1) (see Fig. \ref{fig:case_1}), we present a few examples of comments triggered by its content. The article discusses the Ohio General Assembly's decision to restrict transgender athletes' participation in athletics. The bill limits gender-affirming initiatives for participating in girls' sports. Fig. \ref{fig:case_1} displays comments on the news and user interactions, including responses and counter-responses. Selected examples from this information cascade show comments mentioning entities related to 'sexual orientation' and 'gender'. Notably, no comment includes inter-group mentions, but some comments repeat mentions of the same group. This is evident in the first comment, mentioning both 'trans people' and 'hate women,' both from the 'gender' category. There is no observed intersectionality between categories. The correct and wrong marks indicate whether the tool's detection accurately represents an entity mentioned in a toxic discussion context, meaning both conditions must be met: a mention representing an attack or threat within an aggressive comment. In this article, all entity mentions correspond to mentions of aggressive interactions.

\begin{figure}[h!]
    \centering
    \includegraphics[width=15.3cm]{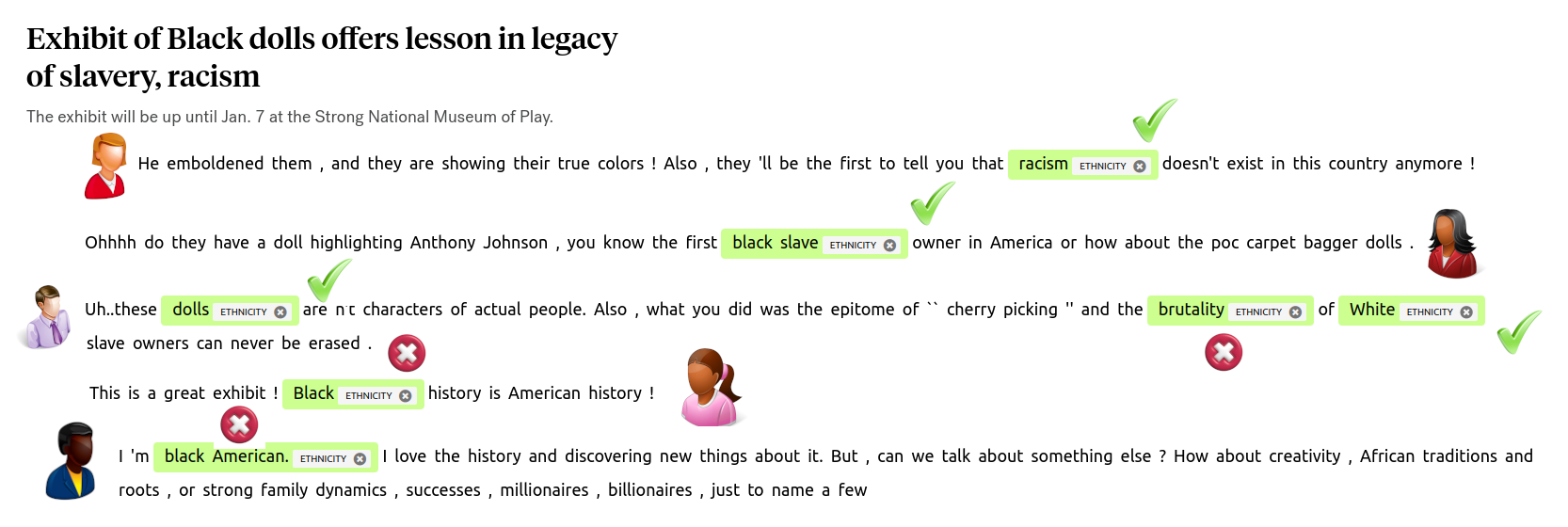}
    \caption{News (7) showing interactions between users. Our tool detects some attacks and mentions regarding racism and ethnicities during the conversation.}
    \label{fig:case_2}
\end{figure}

For news (7) (see Fig. \ref{fig:case_2}), the story covers an exhibition of black dolls, highlighting the presence of slavery in the United States during the 18th century. Such a topic is expected to generate aggressive comments related to 'racism' and 'ethnicities', which is indeed observed in the initial interactions of the example. While the first comment suggests that racism no longer exists, the second delves into details of the exhibition, mentioning Anthony Johnson, an Angolan black man known for acquiring wealth in early 17th-century Virginia, one of the first African American landowners who, through a Virginia court ruling, legally owned a slave. The next three comments feature mentions that do not align with threatened entities. In the third comment, 'brutality' is marked as an entity, a false positive. The same happens with the last comments, where 'black' and 'black American' are tagged as entities from the 'ethnicity' category but not in the context of aggressive comments. Both instances show that these comments are healthy, lacking aggressive expressions, leading to the conclusion that these are false positives: while they tag entities that belong to 'ethnicity', these mentions do not occur in the context of threatening or attacking comments.

\begin{figure}[h!]
    \centering
    \includegraphics[width=16cm]{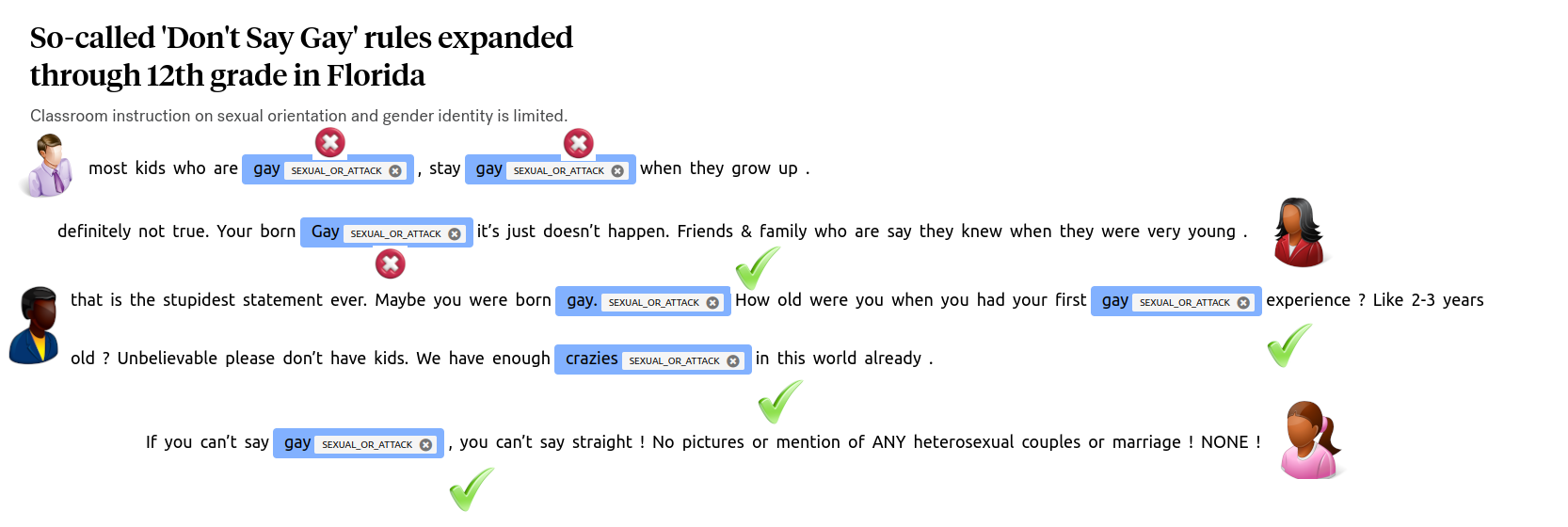}
    \caption{News (15) showing interactions between users. Our tool detects some attacks and mentions regarding sexual orientation during the conversation.}
    \label{fig:case_3}
\end{figure}

The third case, news (15) (see Fig. \ref{fig:case_3}), falls under the 'sexual orientation' category. It addresses the limitations imposed by the 'don't say gay' rules in Florida's school education. A topic like this is expected to stir up conservative and liberal positions regarding 'sexual orientation'. The tool marks mention of entities related to 'sexual orientation,' predominantly the mention of 'gay.' Like the previous case, the first two comments are not aggressive and are marked as false positives, while the latter contain mentions considered threats. These last comments occur in the context of user replies, indicating, as in the previous cases, that aggression arises not concerning the news but from user interactions.

\vspace{3mm}

\noindent \textbf{Main findings}. The tool effectively detects named entities related to the categories of the analyzed news articles. Entities are accurately tagged within their categories, with a negligible error rate for inter-category tagging. However, false positives are observed with tags on tokens not corresponding to entity mentions (e.g., 'brutality'). In some instances, the tool fails to distinguish between aggressive and healthy contexts, marking mentioned entities regardless of the use of toxic language.

\section{Conclusions}
\label{conc}

We have developed a versatile NER tool for identity group tagging, capable of effectively labeling groups across a broad range of categories such as 'gender', 'sexual orientation', 'religion', and 'ethnicity'. This NER was trained by aligning two tasks: entity recognition from the \textbf{HateNorm 2023 dataset}, which identifies text spans of named identity groups, and the \textbf{Jigsaw Toxicity dataset}, created for sentence-level classification. By aligning both datasets, we fine-tuned a conventional NER, enabling it to recognize identity groups in the aforementioned categories. The tool performs well on the test set of the dataset. Subsequently, we applied the tool to social media comments as part of a case study. This study suggests a predominance of comments involving mentions of 'gender' and 'sexual orientation', with a high level of intra-category intersectionality and low inter-category intersectionality. It also reveals a weak correlation between interaction variables and group mentions, with the only significant linear dependence being between 'shares' and mentions of 'sexual orientation'. An analysis of relevant cases shows that the tool effectively detects named entities related to the categories in the analyzed news articles. Entities are accurately tagged within their categories, with a negligible error rate for inter-category tagging. However, false positives occur with tags on tokens not corresponding to entity mentions. In some instances, the tool fails to distinguish between aggressive and healthy contexts, marking mentioned entities regardless of the use of toxic language, which is the main limitation of the tool.

For future work, we aim to refine our NER tool to better differentiate between offensive and neutral mentions of identity groups. This involves expanding our dataset to include more examples from healthy conversations, enhancing the model's ability to distinguish the context in which identity groups are mentioned. Furthermore, we plan to address the complexity of separating offensive mentions directly linked to an attack from those where the identity mention is separate. This distinction will lead us to explore Joint Entity Relation (JER) tagging, leveraging conditional parsing models for a more nuanced analysis. Incorporating advancements in Spanish language models, as seen in the works of Cañete et al. and Araujo et al., will also be crucial in broadening our tool's linguistic and cultural applicability \cite{canete2022albeto,araujo2022evaluation}. 

\vspace{3mm}

\noindent \textbf{Study limitations}. In some instances, the NER tool fails to distinguish between aggressive and healthy contexts, marking mentioned entities regardless of the use of toxic language. Some examples also show that the NER tool produces false positives, mistaking the mention of a group for an attack. This indicates the need for more finely-grained annotations to distinguish between these cases, suggesting using JER-type models to address specific situations. The case study's purpose is exploratory, and no generalizable conclusions can be drawn. It merely represents a case that provides preliminary evidence, requiring further studies to corroborate its findings. The case study does not represent a significant sample of the reality of social networks, as it is limited to only one information source and a single platform. A more comprehensive case study requires the analysis of multiple information sources.
Additionally, comparing different social networks is necessary to compare distinct age groups. We decided not to include additional categories studied in the problem of hate speech detection, such as disability, due to the limited number of annotated examples in this category within the Jigsaw toxicity dataset. Addressing this limitation would require a significant effort in manually annotating examples in a quantity equivalent to that of the datasets employed in this study.

\bibliographystyle{ACM-Reference-Format}
\bibliography{sample-base}

\newpage

\section*{Ethical aspects}
\label{ethics}

\noindent \textbf{Statements on Ethical Considerations}. The study is conducted within the disciplinary line of a publicly funded research center. There has been no interference from supporters in the design or results of the study. The project underwent review by the institutional ethics committee, which approved it without any observations. Data has been securely stored for the purpose of the study and will be deleted after three years. All social media data has been anonymized to protect user privacy. The data has been collected from public sources. Access to META content is under an agreement for the use of the Crowdtangle platform, which allows for the use of its search engine on public pages and the HateNorm and Jigsaw toxicity datasets are public. 

\vspace{3mm}

\noindent \textbf{Statements of Researcher Positionality}. The researchers involved in this study hold personal positions related to their own belief systems, political orientations, and religious beliefs. These views within the group are diverse and have not played a central role in the design and purpose of the study. There are no institutions with political or religious agendas influencing this study. The findings and conclusions of this study have not been subjected to review committees external to the author group, ensuring that they are conveyed without censorship or editorial influence from any group or institution.

\vspace{3mm}

\noindent \textbf{Adverse Impacts Statements}. This research does not present any evident adverse effects. It aims to enhance transparency in online discussions about controversial topics by providing additional resources to researchers and practitioners interested in analyzing mentions of identity groups on social media. The benefits of this study lie in strengthening the management tools of public forums, by gathering information regarding hate speech and the identification of mentions of identity groups on social media.

\end{document}